\documentclass[letterpaper, 10 pt, conference]{ieeeconf}  % Comment this line out if you need a4paper
  \usepackage{pgfplots}
  \pgfplotsset{compat=newest}
  \usepackage{grffile}
  \usepackage{amsmath}
  \usepackage{blindtext}
  \usepackage[bottom]{footmisc}
  \usepackage{mathrsfs}
  \usepackage{lipsum}
 \usepackage{url}
 \usepackage{array}
\usepackage{booktabs}
\usepackage{float}
\usepackage{multirow}

\newlength\fwidth
\usepackage[linesnumbered,ruled]{algorithm2e}
\usepackage{graphicx}
\usepackage{framed}
\usepackage{subcaption}
\usepackage{indentfirst}
\usepackage{multirow}
\usepackage{makecell}

\usepackage{amsmath, amssymb,amsthm}
\usepackage{booktabs}
\usepackage{tabularx}
\usepackage{hyperref}
\usepackage[capitalise]{cleveref}
\usepackage{comment}
\usepackage{romannum}
\usepackage{color}

\IEEEoverridecommandlockouts                              % This command is only needed if 
\title{\LARGE \bf
Efficient Sampling-Based Maximum Entropy Inverse Reinforcement Learning with Application to Autonomous Driving
}

\author{Zheng Wu$^{1,*}$, Liting Sun$^{1,*}$,  Wei Zhan$^{1}$, Chenyu Yang$^{2}$, and Masayoshi Tomizuka$^{1}$% <-this % stops a space
\thanks{*The authors are equally contributed.}% <-this % stops a space
\thanks{$^{1}$Z. Wu, L. Sun, W. Zhan and M. Tomizuka are with the department of Mechanical Engineering, 
        University of California, Berkeley, Berkeley, California, U.S.A.
        Corresponding email: {\tt\small litingsun@berkeley.edu}}%
\thanks{$^{2}$C. Yang is with the Department of Computer Science, Shanghai Jiaotong University, Shanghai, China. The work is conducted during his visit to University of California, Berkeley.}%
}

\begin{document}

\maketitle
\thispagestyle{empty}
\pagestyle{empty}

%%%%%%%%%%%%%%%%%%%%%%%%%%%%%%%%%%%%%%%%%%%%%%%%%%%%%%%%%%%%%%%%%%%%%%%%%%%%%%%%
\begin{abstract}
In the past decades, we have witnessed significant progress in the domain of autonomous driving. Advanced techniques based on optimization and reinforcement learning (RL) become increasingly powerful at solving the forward problem: given designed reward/cost functions, how should we optimize them and obtain driving policies that interact with the environment safely and efficiently. Such progress has raised another equally important question: \emph{what should we optimize}? Instead of manually specifying the reward functions, it is desired that we can extract what human drivers try to optimize from real traffic data and assign that to autonomous vehicles to enable more naturalistic and transparent interaction between humans and intelligent agents. To address this issue, we present an efficient sampling-based maximum-entropy inverse reinforcement learning (IRL) algorithm in this paper. Different from existing IRL algorithms, by introducing an efficient continuous-domain trajectory sampler, the proposed algorithm can directly learn the reward functions in the continuous domain while considering the uncertainties in demonstrated trajectories from human drivers. We evaluate the proposed algorithm on real driving data, including both non-interactive and interactive scenarios. The experimental results show that the proposed algorithm achieves more accurate prediction performance with faster convergence speed and better generalization compared to other baseline IRL algorithms.
\end{abstract}

%%%%%%%%%%%%%%%%%%%%%%%%%%%%%%%%%%%%%%%%%%%%%%%%%%%%%%%%%%%%%%%%%%%%%%%%%%%%%%%%
\section{INTRODUCTION}
\label{intro}
Although rapid progresses have been made in autonomous driving recently, many challenging problems remain open, particularly when interactions between autonomous vehicles (AVs) and humans are considered. Advanced techniques in reinforcement learning (RL) and optimization such as search-based methods\cite{ross2008online}\cite{gu2015tunable}\cite{li2020IFAC}, gradient-based approaches \cite{li2004iterative}\cite{chen2017constrained} and nested optimization \cite{sadigh2018planning} have significantly boosted our capability in finding the optimal trajectories of autonomous vehicles that maximize the specified reward/cost functions. %For instance,  designed a motion planning algorithm which can actively leverage humans' response for more efficient and aggressive maneuvers.
However, to enable more naturalistic and transparent interactions between the AVs and humans, another question is of equal importance: \emph{what should we optimize}? A mis-specified reward function can cause severe consequences. The autonomous vehicles might be either too conservative or too aggressive, neither of which are desirable or safe in real traffic. More importantly, optimizing for reward functions that are misaligned with human expectations will make the behavior of AVs non-transparent to humans, which will eventually degrade the trust from human.

Thus, it is desired to extract what human drivers are optimizing from real traffic data. Inverse reinforcement learning (IRL) \cite{kalman1964linear, ng2000algorithms} have been widely explored and utilized for acquiring reward functions from demonstrations, assuming that the demonstrations are (sub-)optimal solutions of the underlying reward functions. Many examples have proved the effectiveness of such IRL algorithms. For instance, \cite{abbeel2004apprenticeship} proposed an IRL algorithm based on expected feature matching and evaluated it on the control of helicopters. \cite{li2018inverse} proposed an online IRL algorithm to analyze complex human movement and control high-dimensional robot systems. \cite{you2019advanced} used RL and IRL to develop planning algorithm for autonomous cars in traffic and \cite{finn2016guided} designed a deep IOC algorithm based on neural networks and policy optimization, which enabled the PR2 robots to learn dish placement and pouring tasks. In~\cite{sun2018courteous} and \cite{sun_probabilistic_2018}, a courteous cost function and a hierarchical driving cost for autonomous vehicles were learned respectively via IRL to allow autonomous vehicles to generate more human-like behaviors while interacting with humans. \cite{sun2020ICRA} extensively studied the feature selection in IRL for autonomous driving.

Typically, acquiring humans' driving costs from real traffic data has to satisfy a set of requirements:
\begin{itemize}
	\item The trajectories of vehicles are continuous in a high-dimensional and spatiotemporal space, thus the IRL algorithm needs to scale well in high-dimensional continuous space;
	\item The trajectories of the vehicles satisfy the vehicle kinematics and the IRL algorithms should take it into consideration while learning reward functions;
	\item Uncertainties exist in real traffic demonstrations. The demonstrations in naturalistic driving data are not necessarily optimal or near-optimal, and the IRL algorithms should be compatible with such uncertainties;
	\item The features adopted in the reward functions of the IRL algorithms should be highly interpretable and generalizable to improve the transparency and adaptability of the AVs' behaviors.
\end{itemize}

Unfortunately, it is not trivial to satisfy all the above requirements simultaneously. First of all, most existing IRL algorithms, for instance, apprenticeship learning \cite{abbeel2004apprenticeship}, maximum-entropy IRL \cite{ziebart2008maximum} and Bayesian IRL \cite{ramachandran2007bayesian}, are defined within the discrete Markov Decision Process (MDP) framework with relatively small-scale state and action spaces and suffer from the scaling problem in large-scale continuous-domain applications with long horizons. The continuous-domain IOC algorithm proposed in \cite{levine2012continuous} effectively addressed such issue by considering only the local shapes of the reward functions via Laplace approximation. However, it is applicable to deterministic problems where all expert demonstrations are required to be optimal or sub-optimal, which is practically impossible to satisfy via naturalistic driving data, particularly when the features are not known in advance. Moreover, the Laplace approximation requires the calculation of both the gradients and the inverse of the Hessians of the reward functions at the expert demonstrations. When the demonstrations contains long-horizon trajectories, both variables are time-consuming to obtain. The deep IOC algorithm in \cite{finn2016guided} can also work in continuous domain. However, its features are not interpretable, and lead to poor generalization in different scenarios. In \cite{kalakrishnan2013learning}, a sampling-based IRL algorithm was also proposed to address this issue, but it only considered the uncertainties caused by the noises in control and can hardly capture the uncertainties induced by the sub-optimality of different driving maneuvers.

In terms of application domain, our work is closely related to \cite{kuderer2015learning} which also utilize inverse reinforcement learning to learn the reward functions from real driving trajectories. In the forward problem at each iteration, it directly solves the optimization problem and use the optimal trajectories to represent the expected feature counts. However, such optimization process might be quite time-consuming, especially when the driving horizon is long.

Thus, in this paper, we propose an efficient sampling-based maximum entropy IRL (SMIRL) algorithm that satisfies all the above mentioned requirements in autonomous driving. In the algorithm, we explicitly leverage our prior knowledge on efficiently generating feasible long-horizon trajectory samples which allow the autonomous vehicles to interact with the environment, including human drivers. More specifically, in terms of problem formulation, we adopt the principle of maximum entropy. In terms of efficient trajectory sampling for the estimation of the partition term with maximum entropy, we integrate our previous non-conservative and defensive motion planning algorithm with a sampling method to efficiently generate feasible and representative long-horizon trajectory samples. We compare the performance of the proposed SMIRL algorithm with the other two IRL algorithms, i.e., the ones in \cite{levine2012continuous} and \cite{kuderer2015learning} on real human driving data extracted from the INTERACTION dataset. Three sets of evaluation metrics are employed, including both the deterministic metrics such as mean Euclidean distance (MED) and feature count deviation and the probabilistic metric such as the likelihood of the ground-truth trajectories. The experimental results showed that our proposed SMIRL can achieve more accurate prediction performance on the test set in both non-interactive and interactive driving scenarios.

\section{The Method}
\label{our_method}
\subsection{Maximum-entropy IRL}
In this section, we present the proposed SMIRL algorithm to learn reward functions from human driving data.
%\label{our_method:problem_formulation}
% We aim to recover reward function $r$ from driving trajectory set $\{\tau_i\}, i=1,2,...,N$. Such trajectories are represented as the $x$ and $y$ position of the vehicle over time.
Let $x$ denote the states of vehicles, and $u$ denotes the actions. The dynamics of the vehicle, $f(\cdot)$, can then be described as:
\begin{equation}
\label{eqa:dynamic_model}
    x_{k+1} = f(x_k, u_k).
\end{equation}
A driving trajectory in spatial-temporal domain, denoted as $\xi$, contains a sequence of states and actions, i.e., $\xi = [x_0, u_0, x_1, u_1, ..., x_{N-1}, u_{N-1}]$ where $N$ is the length of the planning horizon. Given a set of demonstrations $\Xi_D=\{\xi_i\}$ with $i{=}1,2,\cdots,M$, with the principle of maximum-entropy \cite{ziebart2008maximum}, the IRL problem aims to recover the underlying reward function from which the likelihood of the demonstrations can be maximized, assuming that the trajectories are exponentially more likely when they have higher cumulative rewards (Boltzman noisily-rational model \cite{morgenstern1953theory}):
\begin{equation}
    \label{eq:max_entropy}
    P(\xi, \theta)\propto e^{\beta R(\xi, \theta)}
\end{equation}
where the parameter vector $\theta$ specifies the reward function $R$. $\beta$ is a hyper-parameter that describes how close the demonstrations are to perfect optimizers. As $\beta{\rightarrow}\infty$, the demonstrations approach to perfect optimizers. Without loss of generality, we set that $\beta{=}1$ in this work.

Note that in this work, we assume that all the agents/demonstrations share the same dynamics defined in (\ref{eqa:dynamic_model}). We also assume the reward function underlying the given demonstration set is roughly consistent. Namely, we do not consider scenarios where human drivers change their reward functions along the demonstrations. We also do not specify the diversity of reward functions among different human drivers. Hence, the acquired reward function is essentially an averaged result defined on the demonstration set.

We adopt linear-structured reward function with a selected feature space $\mathbf{f}(\cdot)$ defined over the trajectories $\xi$, i.e.,
\begin{equation}
\label{eqa:reward_function}
    R(\xi, \theta) = \theta^{T}\mathbf{f}(\xi)
\end{equation}
Hence, the probability (likelihood) of the demonstration set becomes
\begin{equation}
\label{eqa:maxentirl_prob}
    P(\Xi_D | \theta) = \prod_{i{=}1}^{M}\frac{e^{\beta R(\xi_i, \theta)}}{{\int_{\tilde{\xi}\in\Phi_{\xi_i}} e^{\beta R(\tilde{\xi}, \theta)} d\tilde{\xi}}} = \prod_{i{=}1}^{M} \frac{1}{Z_{\xi_i}}e^{\beta R(\xi_i, \theta)}
\end{equation}
where $\Phi_{\xi_i}$ represents the space of all trajectories that share the same initial and goal conditions as in $\xi_i$. Our goal is to find the optimal $\theta^*$ which maximizes the averaged log-likelihood of the demonstrations, i.e., 
\begin{equation}
\label{eqa:maxentirl}
    \theta^\star {=} \arg \max_{\theta}\dfrac{1}{M}{\log P(\Xi_{D}| \theta)} {=} \arg \max_{\theta}\dfrac{1}{M}{\sum_{i=1}^{M}\log P(\xi_i|\theta)}.
\end{equation}

From (\ref{eqa:maxentirl_prob}) and (\ref{eqa:maxentirl}), we can see that the key step in solving the optimization problem in (\ref{eqa:maxentirl}) is the calculation of the partition factors $Z_{\xi_i}$. In sampling-based methods, $Z_{\xi_i}$ for each demonstration is approximated via the sum over samples in the sample set $\{\tau^i_m\}, m=1,2,...,K$:
\begin{equation}
\label{eqa:Z_approximation}
% \vspace{-5pt}
    Z_{\xi_i} \approx \sum_{m=1}^{K}{e^{\beta R(\tau^i_m, \theta)}}.
\end{equation}

Thus, the objective function in (\ref{eqa:maxentirl}) becomes:
\begin{align}
    L(\theta) &= \frac{1}{M} \sum_{i=1}^{M}{\log{P(\xi_{i} | \theta)}}\\
    &= \frac{1}{M} \sum_{i=1}^{M}{\log{\frac{e^{\beta R(\xi_i, \theta)}}{\sum_{m=1}^{K}{e^{\beta R(\tau^i_m, \theta)}}}}} \\
    &= \frac{1}{M} \sum_{i=1}^{M}\{\beta R(\xi_i, \theta)-\log{{\sum_{m=1}^{K}{e^{\beta R(\tau^i_m, \theta)}}}}\}.
\end{align}
The derivative is thus given by:
\begin{equation}\label{eqa:maxEntDerivative}
% \vspace{-5pt}
    \nabla_{\theta}L = \frac{\beta}{M}\sum_{i=1}^{M}({\mathbf{f}}(\xi_i) - \widetilde{\mathbf{f}}(\xi_i))
\end{equation}
where 
\begin{equation}
    \widetilde{\mathbf{f}}(\xi_i) = \sum_{m=1}^{K} {\frac{e^{\beta R(\tau^i_m, \theta)}}{\sum_{m=1}^{K}{e^{\beta R(\tau^i_m, \theta)}}} \mathbf{f}(\tau^i_m)}
\end{equation}
defines the expected feature counts over all samples given $\theta$.

Note that an additional $l_1$ regularization over the parameter vector $\theta$ is introduced in the training process to compensate for possible errors induced via the selected set of features.

% $\overline{\mathbf{f}} = \frac{1}{N}\sum_{i=1}^{N}{\mathbf{f}(\xi_i)}$, $\widetilde{\mathbf{f}} = \sum_{m=1}^{M}{\frac{\exp{R(\tau_m, \theta)}}{\sum_{m=1}^{M}{\exp{R(\tau_m, \theta)}}} \mathbf{f}(\tau_m)}$.

% \begin{align*}
%     \begin{equation}
%         L(\theta) &= \sum_{\mathcal{D}}{1} \\
%         &= 1
%     \end{equation}
% \end{align*}

\subsection{The Sampler}
\label{our_method:sampler}
From (\ref{eqa:Z_approximation}), we can see that an efficient sampler is extremely important for solving the aforementioned optimization problem in (\ref{eqa:maxentirl}). Many sampling-based planning algorithms have been widely explored by researchers \cite{karaman_sampling-based_2011, gu2015tunable}. In this section, we integrate the sampler from \cite{gu2015tunable} with our previous work on non-conservative defensive motion planning \cite{zhan2016non} to efficiently generate samples to estimate $Z$ in (\ref{eqa:maxentirl_prob}).

\begin{figure}[h!]
    \centering
    \includegraphics[width=0.8\linewidth]{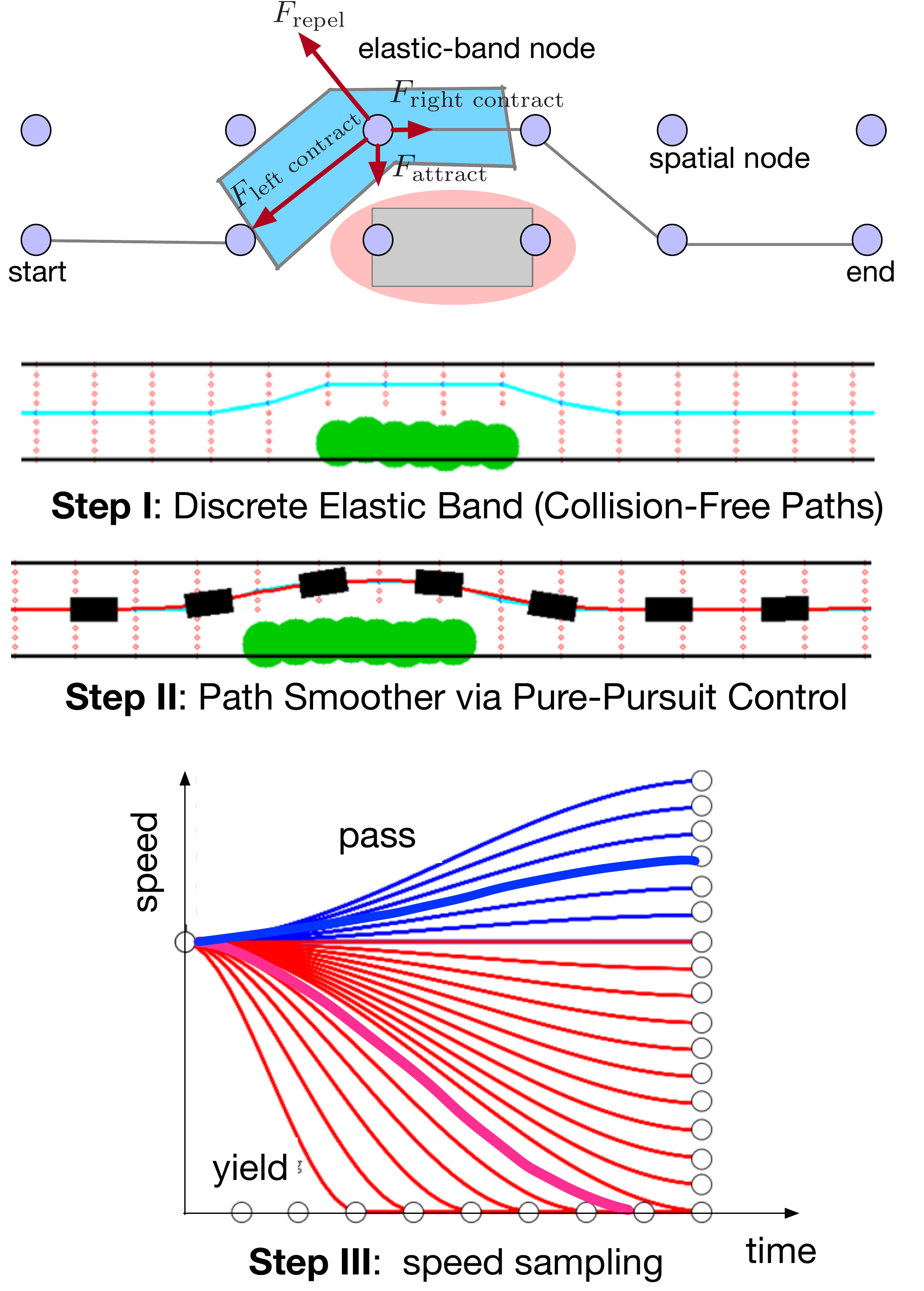}
    \caption{The overview of the sampling process.}
    \label{fig:overview_sampler}
\end{figure}

In \cite{gu2015tunable}, maps are represented via occupancy grids and feasible trajectory samples are generated using decoupled spatiotemporal approaches. As shown in \cref{fig:overview_sampler}, at each time step, the sampler includes three steps: 1) global path sampling via discrete elastic band (ED), 2) path smoothing via pure pursuit control, and 3) speed sampling via optimization and polynomial curves.
%(dual-horizon speed sampling via polynomial curves considering both defensiveness and non-conservativeness of the speed profiles. 

\subsubsection*{Step I - Path Sampling via Discrete ED}
The objective of the first step is to generate collision-free paths. The key insight is that by constraining all samples to be safe, we have intrinsically considered the safety constraints of the optimal problem that the demonstrators try to solve. Moreover, it can also reduce the sample space to approximate $Z$, thus improving the efficiency of the algorithm. To account for moving objects, it considers the sweep-volume of each object of interest within a temporal prediction horizon \cite{gu2015tunable}. As shown in Step I in \cref{fig:overview_sampler}, a path $\tau$ consists of a sequence of elastic nodes (the blue shaded area) which is defined as a spatial node with an IN edge and an OUT edge that connect the neighboring spatial nodes, i.e., $\tau{=}\{\text{node}^i, \text{IN}^i, \text{OUT}^i\}, i{=}1,2,{\cdots}, T.$ For each elastic node, we calculate the weighted sum of three types of forces: the contraction force (both left and right) from neighboring spatial nodes, the repulsive force from obstacles, and the attraction force to drive the vehicle towards desired lane centers. Moreover, collision check is conducted for the IN and OUT edges. Then graph search method is utilized to find a set of collision-free paths that satisfy $\{\mathcal{T}_I\}{=}\{\tau: \text{collision}(\text{IN}){=}0, {\text{Force}(\text{node})}{\le} F_{\text{threshold}}\}$. The $F_{\text{threshold}}$ is a hyper-parameter describing the threshold of the sample set.
\subsubsection*{Step II - Path Smoothing} All the samples in Step I are piece-wise linear but non-smooth paths, which is not feasible for the vehicle kinematics and thus not suitable to estimate $Z$. Hence, in Step II, a pure-pursuit tracking controller is employed to smoothen the paths in $\{\mathcal{T}_I\}$ and generate $\{\mathcal{T}_{II}\}$. One can refer \cite{kuwata2009real} for details.
\subsubsection*{Step III - Speed Sampling} This step will generate speed samples through two components. First, for each path, a suggested speed profile is generated by finding the time-optimal speed plan under physical constraints (e.g., the acceleration/deceleration limits). Second, a local speed curve sampling based on polynomials is introduced to explore the neighborhood of the suggested speed profile. The key insight behind such a two-step speed sampling approach is to reduce the exploration space of the speed profile based on the prior knowledge on human drivers, namely they tend to pursue time-optimal speed plans. To account for discrete driving decisions in interactive scenarios, we will find one suggested speed profile under each decision and conduct separate local speed curve search around them as in \cite{zhan2016non}. For instance, as shown in Step III in \cref{fig:overview_sampler}, when the ego vehicle and another vehicle are driving simultaneously towards an intersection from crossing directions, we will generate suggested speed profile (thick curves in \cref{fig:overview_sampler}) and local speed samples under both the ``yield'' (red) and the ``pass'' (blue) decisions. For the speed curve, we use third-order polynominals to generate samples, as addressed in \cite{gu2015tunable}. Hence, via Step III, a spatiotemporal trajectory set $\{\mathcal{T}_{III}\}$ is generated. 

\subsection{Re-Distribution of Samples}
\label{our_method:re-distribution}
\begin{figure}[h!]
    \centering
    \includegraphics[width=0.9\linewidth]{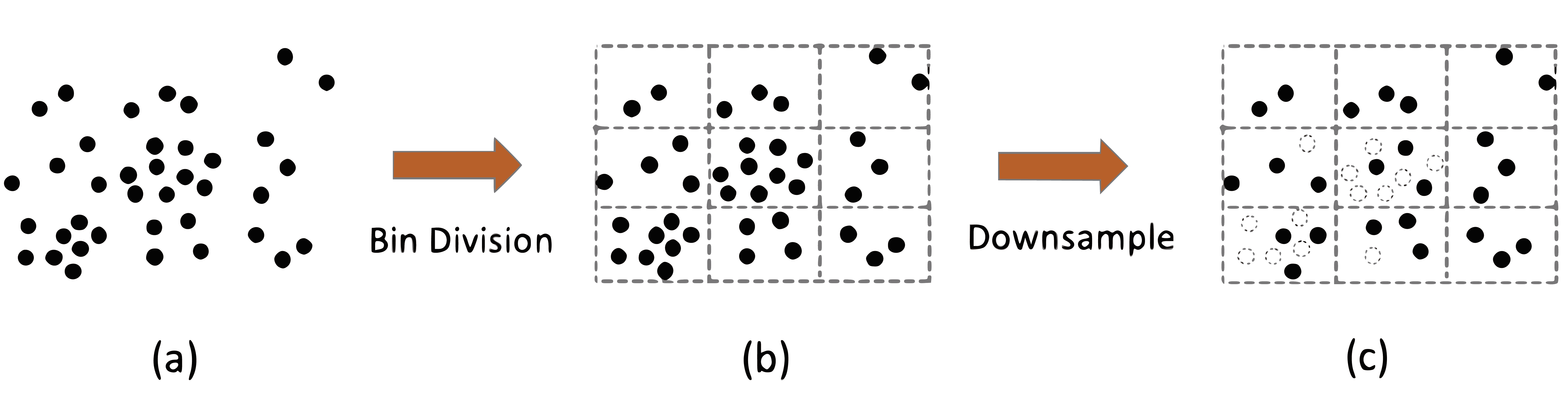}
    \caption{Re-distribution of samples}
    \label{fig:sampling_bin}
\end{figure}

Note that in (\ref{eqa:maxentirl_prob}), the probability of a trajectory $\xi$ is evaluated via the normalization term $Z$ which is estimated via the samples in $\{\mathcal{T}_{\Romannum{3}}\}$. However, the samples in $\{\mathcal{T}_{III}\}$ are not necessarily uniformly distributed in the selected feature space $\mathbf{f}(\xi)$, which will cause biased evaluation of probabilities, as pointed in \cite{bobu2020less}. To address this problem, we propose to use Euclidean distance in the feature space as a similarity metric for re-distributing the samples. As shown in \cref{fig:sampling_bin}, the re-distribution process takes two steps: 1) with an initial sample set $\{\mathcal{T}_{III}\}$ (\cite{bobu2020less}(a)), we uniformly divide the feature space into discrete bins, and 2) re-sample within each bin to make sure that each bin has roughly equal number of samples.

\subsection{Summary of the SMIRL algorithm}
\label{our_method:overview}
The proposed SMIRL algorithm can thus be summarized in Algorithm \ref{algorithm_SMIRL}.
% \vspace{-5pt}
\begin{algorithm}[h!]
\caption{The Proposed Sampling-based Maximum Entropy IRL for Driving\label{algorithm_SMIRL}
}
\SetAlgoLined
\KwResult{optimized reward function parameters $\theta^\star$}
 \KwIn{The demonstration dataset $\mathcal{D}_M = \{\xi_i\}_{i=1:M}$, the convergence threshold $\epsilon$ and the learning rate $\alpha$.}
 Initialize $\theta_0$, $k=0$ and compute expected expert feature count $\overline{\mathbf{f}}(\mathcal{D}_M)=\frac{1}{M}\sum_{i=1}^{M}{\mathbf{f}(\xi_i)}$\;
 
 Generate the sample set $\mathcal{D}^0_s = \{\tau^i_m\}_{m=1:K, i=1:M}$ using the sampler in \Cref{our_method:sampler}\;
 
 Re-distribute the samples according to their similarities as discussed in \Cref{our_method:re-distribution}, and generate a new sample set $\mathcal{D}_s$\;
 
 Compute the initial expected feature count over all samples $\widetilde{\mathbf{f}}_0(\mathcal{D}_s) = \dfrac{1}{M}\sum_{i=1}^{M}\widetilde{\mathbf{f}}_0(\xi_i) = \dfrac{1}{M}\sum_{i=1}^{M}\frac{1}{K}\sum_{m=1}^{K}{\frac{\exp{R(\tau^i_m, \theta_0)}}{\sum_{m=1}^{M}{\exp{R(\tau^i_m, \theta_0)}}} \mathbf{f}(\tau^i_m)}$\;
 
 \While{$\Vert\overline{\mathbf{f}}(\mathcal{D}_M)-\widetilde{\mathbf{f}}_k(\mathcal{D}_s)\Vert_2\ge\epsilon$} 
  {
  	%Compute reward $R(\tau, \theta_k)$ for each sample in $\mathcal{D}_s$\;
	Update $\theta_k$ using gradient decent, i.e., $\theta_{k+1} = \theta_k + \nabla_{\theta_k}L = \theta_k +  \alpha(\overline{\mathbf{f}}(\mathcal{D}_M)-\widetilde{\mathbf{f}}(\mathcal{D}_s))$\;
  	Compute the expected feature count based on $\theta_{k+1}$ over all samples $\widetilde{\mathbf{f}}_{k+1}(\mathcal{D}_s) = \dfrac{1}{M}\sum_{i=1}^{M}\widetilde{\mathbf{f}}_{k+1}(\xi_i) = \dfrac{1}{M}\sum_{i=1}^{M}\frac{1}{K}\sum_{m=1}^{K}{\frac{\exp{R(\tau^i_m, \theta_{k+1})}}{\sum_{m=1}^{M}{\exp{R(\tau^i_m, \theta_{k+1})}}} \mathbf{f}(\tau^i_m)}$\;
	$k = k + 1$\;
  }
 $\theta^* = \theta_k$\;
\end{algorithm}
% 	\vspace{-10pt}
	
\section{Experiments on Driving Behavior}
\label{experiment}
We apply the proposed SMIRL to learn the human driving behavior from real traffic data. Experiments on two types of driving scenarios are conducted: one focusing on independent driving behavior (i.e., non-interactive driving (NID)) and the other on interactive driving (ID) behavior at merging traffic. The NID scenario aims to recover humans' preference when they are driving freely, and the ID case aims to capture how human drivers interact with others in merging traffic. 

% \vspace{-5pt}
\subsection{Dataset}
\label{experiment:dataset}
\begin{figure}[h]
%   \centerin the roadng
  \begin{subfigure}[t]{0.24\textwidth}
        \centering
        \includegraphics[width=\textwidth]{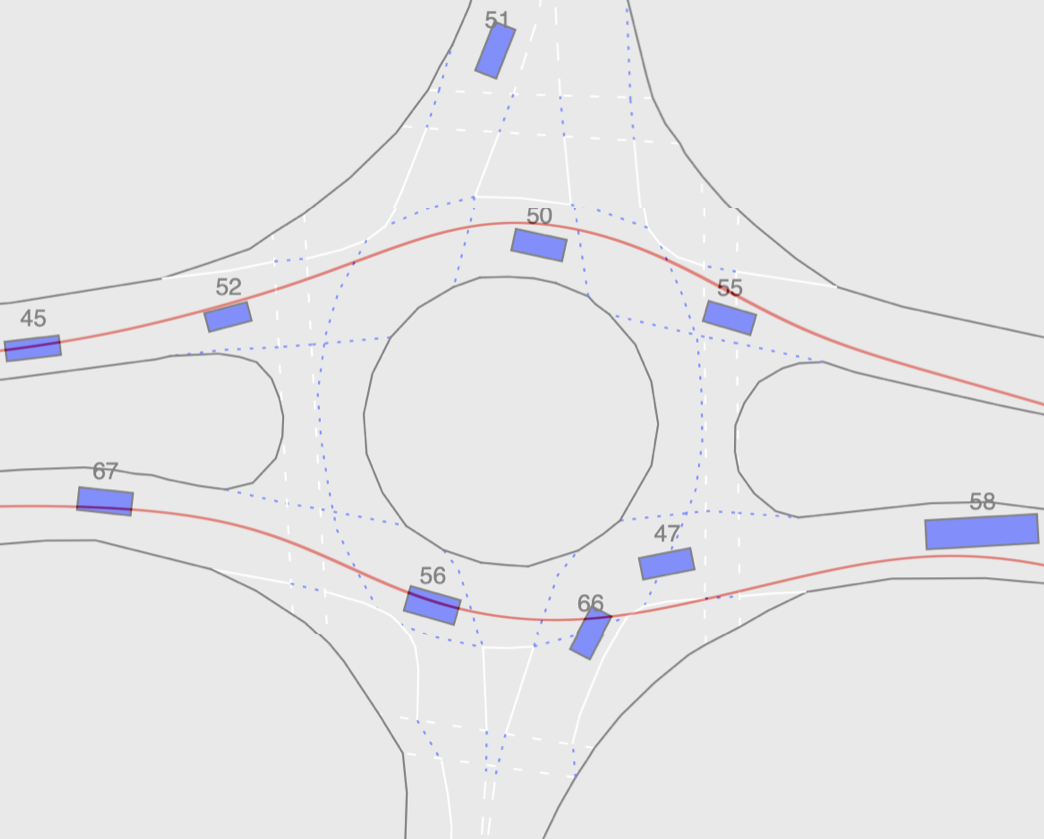}
        \caption{The NID scenario with DR\_USA\_Roundabout\_SR}
    \end{subfigure}%
    ~ 
    \begin{subfigure}[t]{0.24\textwidth}
        \centering
        \includegraphics[width=\textwidth] {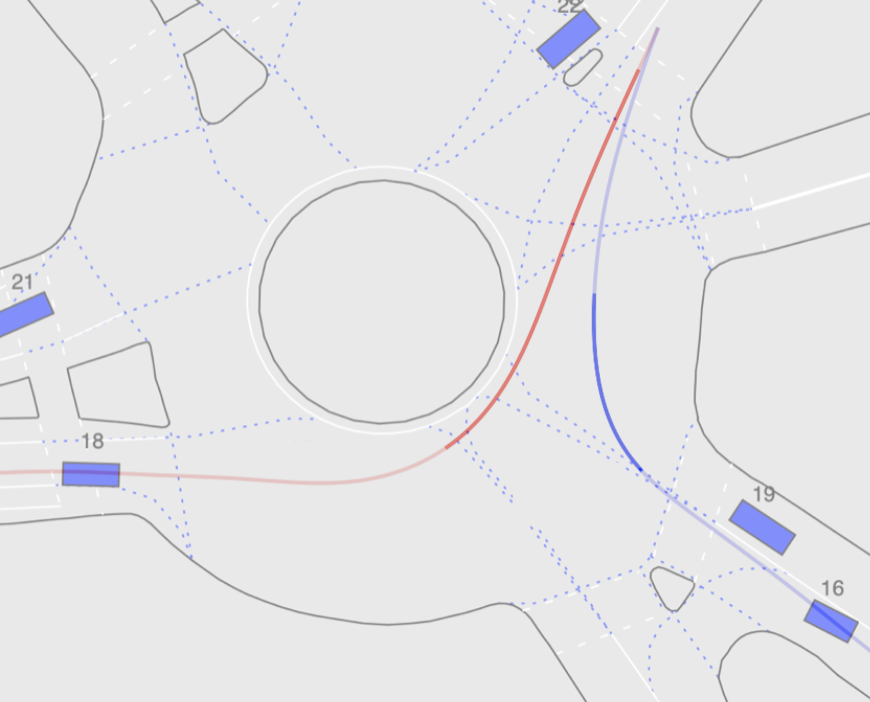}
        \caption{The ID scenario with DR\_USA\_Roundabout\_FT}
    \end{subfigure}
% 	\vspace{-5pt}
%   	\subfloat[Non-interactive ]{\includegraphics[width=0.15\textwidth]{./pics/SR_R.png}}\quad
% 	\subfloat[Interactive]{\includegraphics[width=0.15\textwidth]{./pics/FT_R.png}}
 \caption{Two roundabout scenarios.\label{fig:data_sample}}
\end{figure}

We select training data from the INTERACTION dataset~\cite{zhan2019interaction, zhan2019constructing} with the NID trajectories from the subset DR\_USA\_Roundabout\_SR and ID trajectories from the subset DR\_USA\_Roundabout\_FT, as shown in \cref{fig:data_sample}.

In the NID scenario, we select 113 driving trajectories which travel across the roundabout horizontally, as demonstrated by the red lines in \cref{fig:data_sample}(a). 80 trajectories are used as training data and the remaining 33 as test data. In the ID scenario, we selected 233 pairs of interactive driving trajectories with two vehicles: an ego vehicle trying to merge into the roundabout and an interacting vehicle that is already in the roundabout (interactive trajectories at different locations of the roundabout are contained). An illustrative example is shown in \cref{fig:data_sample}(b) where the blue and red lines represent, respectively, the trajectories of the ego vehicle and the interacting vehicle. The solid segments of the red and blue lines in \cref{fig:data_sample}(b) indicate the time period when the two vehicles are interacting with each other, and we use them for learning. 150 pairs of trajectories are for training and the other 83 for testing. The sampling time of all trajectories is $\Delta t = 0.1s$. 

% \vspace{-5pt}

% \vspace{-15pt}
\subsection{Feature Selection}
\label{experiment:feature_selection}
Recall that in (\ref{eqa:reward_function}), we assume that the reward function is a linear combination of a set of specified features. The goal of such a feature space is to explicitly capture the properties that humans care when they are driving. We categorize the features into two types: non-interactive features and interactive features.
\subsubsection*{Non-interactive features}
\paragraph*{Speed} To describe human's incentives to drive fast and the influence of traffic rules, we define the speed feature as
%\vspace{-8pt}
\begin{equation}
    f_v(\xi) = \frac{1}{N}\sum_{i=1}^{N}{(v_i - v_{\text{desired}})^2}
\end{equation}
where $v_{\text{desired}}$ is the speed limit.

\paragraph*{Longitudinal and lateral accelerations} Accelerations are related to not only the power consumption of the vehicle, but also to the comfort of the drivers. To learn human's preference on them, we include both of them as features:
\begin{eqnarray}
    f_{a\_lon}(\xi) &= &\frac{1}{N}\sum_{i=1}^{N}{a_{lon}^2}\\
    f_{a\_lat}(\xi) &= &\frac{1}{N}\sum_{i=1}^{N}{a_{lat}^2}
\end{eqnarray}
\paragraph*{Longitudinal jerk} We also have included longitudinal jerk as a feature since it can describe the comfort level of human's driving behavior. It is defined as:
\begin{equation}
    f_{jerk}(\xi) = \frac{1}{N}\sum_{i=1}^{N}{(\frac{a_t - a_{t-1}}{\Delta t})^2}
\end{equation} 
\subsubsection*{Interactive features} 
\begin{figure}[h!]
\centering
\includegraphics[width=0.88\linewidth]{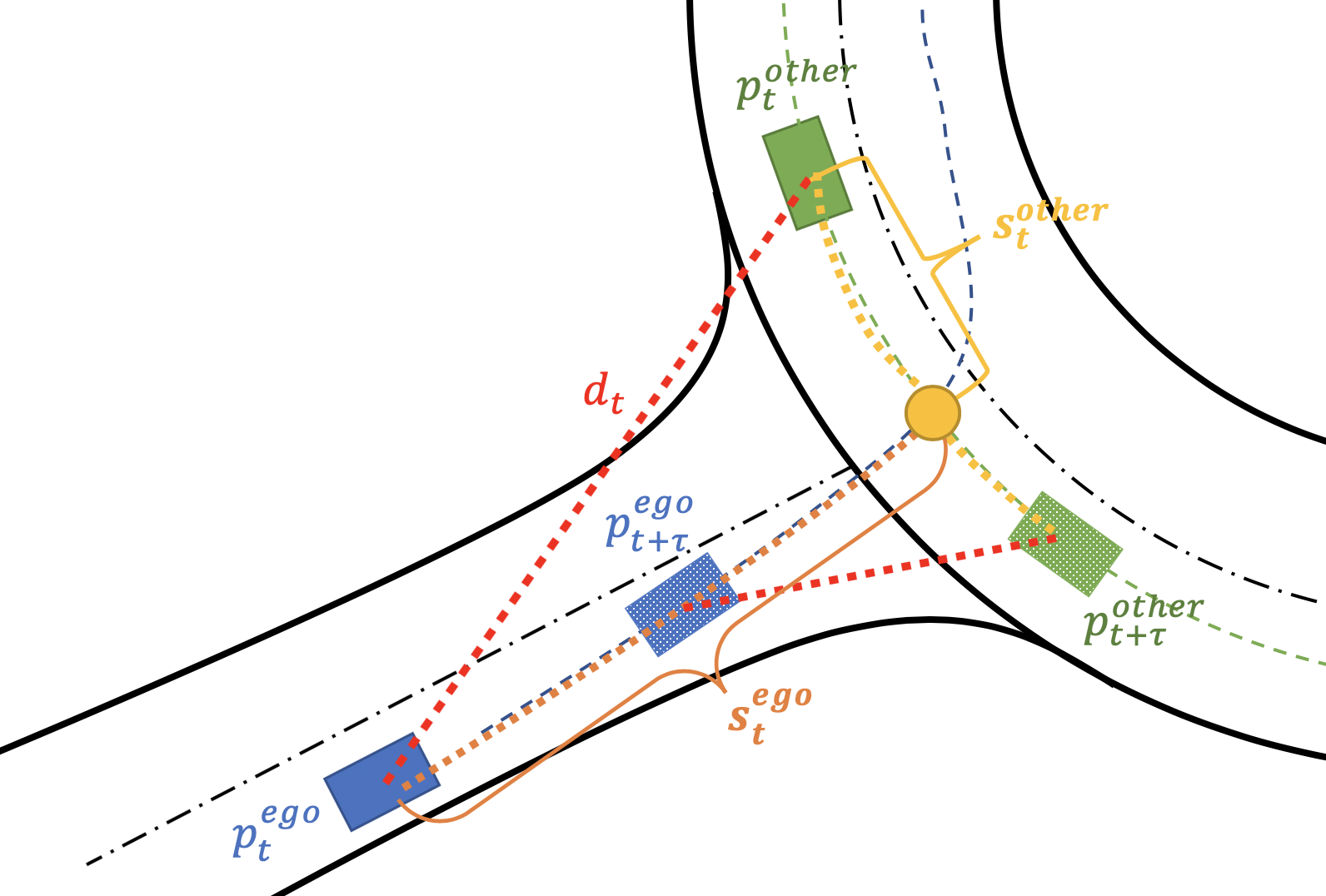}
\caption{The illustration of the proposed interactive features. The blue and green rectangles represent the ego and the other vehicle, respectively. The orange circle is the conflict point of the two vehicles on their routes. The spatial distance $d$ at $t$ is demonstrated via red dotted lines. The longitudinal distances to the collision point for both vehicles at $t$ are shown by the yellow dot curves.}
\label{fig:FtrDefn}
% \vspace{-10pt}
\end{figure}

To capture the mutual influence between interactive drivers, we define two interactive features.
\paragraph*{Future distance} The future distance $d$ is defined as the minimum spatial distance of two interactive vehicles within a predicted horizon $\tau_{\text{predict}}$ assuming that they are maintaining their current speeds (in this paper, we use $\tau_{\text{predict}}{=}1s$). As demonstrated in \cref{fig:FtrDefn}, between time $t$ and $t{+}\tau$, the blue vehicle drives from $p^{\text{ego}}_t$ to $p^{\text{ego}}_{t{+}\tau}$ and the green one drives from $p^{\text{other}}_t$ to $p^{\text{other}}_{t{+}\tau}$. At time $t$, their spatial distance is demonstrated via $d_t$. Hence, The feature of future distance $d$ is given by
% \vspace{-3pt}
\begin{equation}
    f_{dist}(\xi ) = \frac{1}{N} \sum_{i=1}^{N} e^ { - \min _{\tau \in [0,\tau_{\text{predict}}]} d\big(t_i+\tau \big)}.
\end{equation}
\paragraph*{Future interaction distance}
Different from $f_{dist}(\xi )$, the feature related to future interaction distance is defined as the minimum distance between their distances to the collision point, i.e., 
% \vspace{-3pt}
\begin{equation}
f_{\text{int\_dist}}(\xi ) = \frac{1}{N} \sum_{i=1}^{N}e^{ -\min_{\tau \in [0,\tau_{\text{predict}}]} \vert s^{\text{ego}}(t_i+\tau)-s^{\text{other}}(t_i+\tau)\vert}
\end{equation}
where $s^{\text{ego}}(t_i+\tau)$ and $s^{\text{other}}(t_i+\tau)$ represent, respectively, the longitudinal distances of the blue and green vehicles to the shared collision point (the orange circle in \cref{fig:FtrDefn}) at time $t_i{+}\tau$. Again, we assume the vehicles maintain their speeds at $t_i$ through the horizon $\tau_{\text{predict}}$.

Note that the above features all have different physical meanings and units. Hence, to assure fair comparison of their contributions to the reward function, we normalize all of them to be within (0,1) before the learning process by dividing them by their own maximum values on the dataset.

% \adnote{add chenyu's plot here.}
%%% ycy

% \vspace{-5pt}
\subsection{Baseline Methods}
\label{experiment:baselines}
To validate the effectiveness and efficiency of our proposed method, we compare our method with three other representative IRL algorithms: the continuous-domain IRL ({CIOC}) in \cite{levine2012continuous}, the optimization-approximated IRL ({Opt-IRL}) in \cite{kuderer2015learning}, and the guided cost learning (GCL) algorithm in \cite{finn2016guided}. CIOC, Opt-IRL and our method are model-based, while GCL is model-free (i.e., deep learning based). All of them are based on the principle of maximum entropy, but differ in the estimation of $Z$.
\begin{itemize}
    \item {CIOC} estimates $Z$ in a continuous domain via Laplace approximation. Specifically, the reward at an arbitrary trajectory $\tilde{\xi}$ can be approximated by its second-order Taylor expansion at a demonstration trajectory $\hat{\xi}_{D}$, i.e., 
    \begin{equation}
    \label{eqa:laplace_approximation}
    \begin{aligned} 
    R(\theta, \tilde{\xi}) \approx R(\theta, \hat{\xi}_{D} )+(\tilde{\xi}-\hat{\xi}_{D})^{T} \frac{\partial R}{\partial \xi_{D}} \\+(\tilde{\xi}-\hat{\xi}_{D})^{T} \frac{\partial^{2} R}{\partial \xi_{D}^{2}}(\tilde{\xi}-\hat{\xi}_{D}).
    \end{aligned}
    \end{equation}
    This simplifies $Z{=}\int_{\tilde{\xi}}e^{\beta R(\theta, \tilde{\xi})}d\tilde{\xi}$ as a Gaussian integral which can be directly calculated. For more details, one can refer to \cite{levine2012continuous}.
    
    \item {Opt-IRL} estimates $Z{=}\int_{\tilde{\xi}}e^{\beta R(\theta, \tilde{\xi})}d\tilde{\xi}$ via the optimal trajectory $\xi_{opt}$. Namely at each training iteration, with the updated $\theta$, an optimal trajectory $\xi_{opt}$ can be obtained by minimizing the updated reward function, and $Z{\approx}e^{\beta R(\theta, \xi_{opt})}$ is utilized as an approximation. %to re-generate trajectory $\tilde{\xi}_i$ given the initial state of $\xi_i$ for each trajectory $\xi_i$ in the training set $\mathcal{D}$. Then they use $\tilde{\mathcal{D}} = \{\tilde{\xi_i\}}_{i=1,...,N}$ to estimate the partition function $Z$, $Z \approx \sum_{\tilde{\mathcal{D}}}\exp{R(\tilde{\xi_i}, \theta)}$.
    
    \item Different from model-based IRL, GCL parameterizes the reward function as well as the policy via two deep neural networks which are trained together. It uses rollouts (samples) of the policy network to estimate $Z$ in each iteration. Note that the key difference between GCL and the model-based IRL is that GCL does not need manually crafted features, but automatically learns features via the neural networks.
    
\end{itemize}

% \cite{levine2012continuous} used Laplace approximation to estimate the partition function $Z$. Specifically, [ZHENG: how to make this more detailed?]

% \cite{kuderer2015learning} adopted traditional IRL framework by solving a forward optimization problem in each training iteration. Specifically, in each training iteration, we use current $\theta$, i.e., current reward function, to re-generate trajectories given the initial state for each trajectory. Then we use those re-generated trajectories to estimate the partition function $Z$, $Z \approx $

%While in our method, we utilize the prior knowledge from driving and propose an efficient sampling-based method to estimate the partition function $Z$. For convenience, we refer \cite{levine2012continuous} as \textbf{CIOC} and \cite{kuderer2015learning} as \textbf{Opt-IRL}. Our method is referred as \textbf{Ours}. We compare our sampling-based method with the two methods and show these results in Section~\ref{experiment:result}

\subsection{Evaluation Metrics}
\label{experiment:evaluation}
We employ three metrics to evaluate the performance of different IRL algorithms: 1) feature deviation from the ground truth as in \cite{abbeel2004apprenticeship}, 2) mean Euclidean distance to the ground truth as in \cite{sun2018courteous} and \cite{sun_probabilistic_2018}, and 3) the likelihood of the ground truth. Definitions of the three metrics are given below.
\subsubsection{Feature Deviation}
%The essence of IRL lies at aiming to find the optimal reward function that makes the re-generated trajectories match the ground truth trajectories best. Thus we use 
Given a learned reward function, we can correspondingly generate a best predicted future trajectory for every sample in the test set. %of the demonstration trajectories in the test set, we will evaluate the probabilities of all samples and pick the one with highest probability as the predicted trajectory. 
The feature deviation (FD) between the predicted trajectories and the ground-truth trajectories is defined as follows:
% \vspace{-3pt}
\begin{equation}
\mathcal{E}_{FD} = \frac{1}{M}\sum^{M}_{i=1}\frac{1}{N_i}{\frac{\vert \mathbf{f}(\xi_i^{gt}) - \mathbf{f}(\xi_i^{pred}) \vert}{\mathbf{f}(\xi_i^{gt})}}.
\end{equation}
where $M$ is the number of trajectories in the test set, and $N_i$ is the length of the $i$-th trajectory.
% We compare the relative error of the feature count between the predicted trajectories and their corresponding ground truth trajectories. %ycy
% \begin{equation}
% \mathcal{E}_{FD} = \frac{1}{M}\sum^{M}_{i=1}\left\vert f(\xi_i^\text{gt}) - f(\xi_i^\text{gt}) \right\vert
% \end{equation}
\subsubsection{Mean Euclidean Distance (MED)}
The mean Euclidean distance is defined as:
\begin{equation}
\mathcal{E}_{MED} = \frac{1}{M}\sum^{M}_{i=1}{\frac{1}{N_i} ||\xi_i^{gt} - \xi_i^{pred}||_2}.
\end{equation}

%[YCY: trajectory's defn is a sequence of tuple (x and u), dose not mean spacial position]
%[YCY: $\xi$ is not indexed by i?]
\subsubsection{Probabilistic Metrics}
We also evaluate the likelihood of the ground-truth trajectories given the learned reward functions through different IRL algorithms. Recalling (\ref{eqa:maxentirl_prob}), the likelihood of a ground-truth demonstration $\xi$ in the test set is given by
\begin{equation}
    P(\xi | \theta, \{\mathcal{T}\}) = \frac{\exp{(R(\xi, \theta))}}{\exp{(R(\xi, \theta))} + \sum_{i=1}^{M}{\exp{(R(\tau_i, \theta))}}},
\end{equation}
where $\{\mathcal{T}\}$ is the set of samples generated via our proposed approach. In our proposed method, CIOC and Opt-IRL, $R(\xi,\theta){=}\theta^T \mathbf{f}(\xi)$, and in GCL, it is given via a neural network. Among the four IRL algorithms, the one that generates the highest likelihood on the ground-truth trajectories wins.

\subsection{Experiment Conditions}
Among the four different IRL algorithms (i.e., ours vs the three baselines), the CIOC, Opt-IRL and ours are model-based and need manually selected features. GCL, on the other hand, learns to extract features. Hence, for first three approaches, we used the four mentioned features in the non-interactive case (\textit{speed} (v\_des), \textit{longitudinal acceleration} (a\_lon), \textit{lateral acceleration} (a\_lat) and \textit{longitudinal jerk} (j\_lon)) and six features in the interactive case (\textit{speed} (v\_des), \textit{longitudinal acceleration} (a\_lon), \textit{lateral acceleration} (a\_lat), \textit{longitudinal jerk} (j\_lon), \textit{future distance} (fut\_dis), and \textit{future interaction distance} (fut\_int\_dis)). For the GCL algorithm, no manual features were utilized, and the inputs were directly trajectories.

Other hyper-parameters of the proposed algorithm is set as follows: $F_{\text{threshold}}=1.0$ with weights on different forces as $w_{\text{contract}}=w_{\text{repel}}=w_{\text{attract}}=1/3$, i.e., no preference was introduced during the path sampling process over the smoothness of the paths, deviations from the reference lane, and the distance to obstacles.

%For each trajectory in test set, we compute the probability of the ground truth trajectories given $\theta$ trained with different IRL algorithms. The joint probability of all trajectories in the test set is computed for comparison. The sample sets are kept the same across different algorithms.

    %\item Win Count. Similar to Trajectory Joint Probability, we compute the probability of ground truth trajectories given $\theta$ trained with different IRL algorithms. We record the number of trajectories whose probability is larger than the other two methods as the Win Count of certain method.
\section{Results and Discussion}
\label{experiment:result}
The performances of the four algorithms, i.e., ours, CIOC, Opt-IRL and GCL, are evaluated based on the metrics in \cref{experiment:evaluation}. We conducted two types of tests: one on test sets in the same environment (seen) as the training sets, and one on completely new test sets in a new merging environment (unseen).

\subsection{Performance on Test Sets in Seen Environments}
\label{subsection:seen_tests}
The results of the four IRL algorithms under the non-interactive and interactive driving scenarios are listed in \cref{table:noninteractive} and \cref{table:interactive}, respectively. 
%we use Euclidean distance (ED) and feature count deviation (FD) as our evaluation metrics. During training, we apply three different IRL algorithms, i.e., \textbf{Ours}, \textbf{CIOC} in \cite{levine2012continuous}, \textbf{Opt-IRL} in \cite{}, to learn the corresponding reward function from the demonstrations in the training set. In the test phase, [ZHENG: add test description, optimization-based and sampling-based separately] we apply The results are shown in Table~\ref{table:noninteractive}

\begin{table*}
\centering
\caption{A summary of the IRL algorithms in the non-interactive driving scenario.}
\label{table:noninteractive}
\begin{tabular}{c|cccc|c|c|c}
\hline
    &  a\_lon &  j\_lon &  v\_des &  a\_lat &  MED & Win Count & Log Likelihood \\
\hline
\textbf{Ours} &      0.16$\pm$0.12 &      0.20$\pm$0.15 &      0.09$\pm$0.04 &      0.09$\pm$         0.03 &     0.21$\pm$        0.06 &33 & -238.98 \\
\hline
    \textbf{Opt-IRL} &      0.19$\pm$         0.19 &      0.32$\pm$         0.19 &      0.13$\pm$         0.06 &      0.11$\pm$         0.03 &     0.29$\pm$        0.09 & 0 & -398.93 \\
\hline
  \textbf{CIOC} &      0.48$\pm$         0.42 &      0.23$\pm$         0.17 &      0.10$\pm$         0.07 &      0.06$\pm$         0.05 &     0.23$\pm$        0.09 & 0 & -662.16 \\
\hline
\textbf{GCL} &     --- &  ---  &  --- &    ---  &     3.73$\pm$ 1.95 & 0 & -1377.65 \\
\hline
\end{tabular}
\end{table*}

\begin{table*}
\centering
\caption{A summary of the IRL algorithms in the interactive driving scenario.}
\label{table:interactive}
 \begin{tabular}{c|cccccc|c|c|c}
\hline
    &  a\_lon &  j\_lon &  a\_lat &  v\_des &  fut\_dis &  fut\_int\_dis &   MED & Win Count & Log Likelihood\\
\hline
\textbf{Ours} &      $\makecell{0.15\pm \\ 0.24}$ &     $\makecell{0.54\pm \\ 0.19}$ &      $\makecell{0.19\pm \\ 0.24}$ &  $\makecell{0.034\pm \\ 0.026}$ &                $\makecell{0.012\pm \\ 0.0078}$ &  $\makecell{0.032\pm \\ 0.045}$  &     $\makecell{0.066\pm \\ 0.038}$ & 63 & -515.97\\
\hline
\textbf{Opt-IRL} &      \makecell{0.69$\pm$ \\ 1.04} &      \makecell{0.55$\pm$     \\    0.40} &     \makecell{ 0.20$\pm$   \\      0.23} &    \makecell{      0.083$\pm$    \\         0.11} &      \makecell{  0.021$\pm$  \\                 0.018} &   \makecell{ 0.043$\pm$   \\      0.066} &    \makecell{ 0.14$\pm$  \\      0.16} & 4 & -802.01 \\
\hline
 \textbf{CIOC} &    \makecell{  0.42$\pm$    \\     0.77} &     \makecell{ 0.69$\pm$  \\       0.26} &   \makecell{   0.26$\pm$   \\      0.23} &         \makecell{ 0.064 $\pm$     \\  0.10} &       \makecell{  0.023$\pm$ \\                  0.012} &     \makecell{  0.045$\pm$     \\     0.10} &     \makecell{0.14$\pm$   \\  0.14} & 9 & -595.27 \\
\hline
\textbf{GCL} &  --- &  ---  &  --- &    --- &  ---  &   ---  &     \makecell{1.53$\pm$   \\  1.16} & 0 &  -1196.75 \\
\hline
\end{tabular}
\end{table*}
\begin{table}[h!]
	\centering
	\caption{The learned weights of reward function using our proposed approach (NID refers to non-interactive driving scenario and ID refers to interactive driving scenario)}
	\label{table:Weights}
	\begin{tabular}{c|cccccc}
		\hline
        &   a\_lon &   j\_lon &   a\_lat &   v\_des &   fut\_dis &   fut\_int\_dis \\
        \hline
NID & 1 &   0.363      &     0.001     &  0.14   & - & -\\
ID & 1 &    0.007   &    0.002   &     0.102      & 0.007& 0.022       \\
		\hline
	\end{tabular}
\end{table}

\subsubsection{Feature Deviation}
Feature deviation was evaluated among the three model-based approaches, i.e., ours, CIOC and Opt-IRL. We can see that compared to the other two algorithms, the proposed SMIRL algorithm achieved relatively smaller $\mathcal{E}_{FD}$ on most of the features in non-interactive scenario, except for the feature $a\_lat$. In the interactive scenario, the proposed algorithm achieved smaller $\mathcal{E}_{FD}$ on all features. Moreover, the variations of $\mathcal{E}_{FD}$ are consistently smaller across different features and scenarios. This means that the proposed algorithm learned a reward function that can better capture the general driving preference in the dataset, and thus achieve more stable performance. The learned weights via our proposed approach are given in \cref{table:Weights}. We can see that the contribution of the feature $a\_lat$ is relatively small in both scenarios, particularly in the non-interactive case. Therefore, the proposed algorithm generated a relatively large feature deviation on $a\_lat$ in the non-interactive case.

\subsubsection{MED}
MED was evaluated among all four IRL algorithms. In \cref{table:noninteractive} and \cref{table:interactive}, we can see that our method achieved smaller MEDs and variances compared to the {CIOC}, {Opt-IRL} and {GCL} in both driving scenarios. This indicates that the our algorithm can find a reward function that better explains human driving behaviors, i.e., generates more {human-like} trajectories. An interesting finding is that GCL, as a deep learning based method, did not necessarily achieve better performance than the model-based IRL algorithms. There might be multiple reasons leading to such an outcome. First, the training sets we utilized were too small for deep learning based models to converge well. Second, the data from real traffic contained noises which might deteriorate the performance of GCL, particularly when the amount of data was not sufficient. Such outcomes also further verified the data efficiency and better robustness of model-based IRL algorithms in the presence of data noises. 

\subsubsection{Probabilistic Metric}
The results of the four methods in terms of the probabilistic metric are also shown in \cref{table:noninteractive} and \cref{table:interactive}. We can see that in both scenarios, the ground-truth trajectories in the test sets all have higher likelihood using the reward function retrieved by our approach compared to those by the other three algorithms, which demonstrates the effectiveness of our proposed approach.

The learned weights in \cref{table:Weights} indicate that humans care more about longitudinal accelerations in both non-interactive and interactive scenarios. During interaction, human drivers pay more attention to future interaction distance, and less on longitudinal jerks and lateral accelerations. 

\subsection{Performance on Test Sets in Unseen Environments}
\label{subsection:unseen_tests}
To validate the robustness and generalization ability of our proposed method, we also conducted a comparison among the four IRL algorithms on new test sets in an unseen environment in the training sets. More specifically, we trained all four algorithms using the roundabout merging in the training data and directly tested the trained models on other unseen merging scenarios with different road structures. Both interactive and non-interactive scenarios were tested. The results were given in \cref{table:generalization_deterministic} and \cref{table:generalization_prob}. It is shown that all model-based IRL algorithms, including ours, can generalize better compared to GCL. Both the MED and likelihood did not degrade as significantly as the GCL algorithm.
\begin{table}[h]
	\centering
	\caption{Generalization results of different IRL algorithms under the MED metric. The results are in meters.}
	\label{table:generalization_deterministic}
	\begin{tabular}{c|cc| cc}
		\hline
		&  {Seen NID} &  {Unseen NID} &  {Seen ID} &  {Unseen ID}\\
		\hline
		Ours &  0.21  &  0.74  & 0.066 & 0.072 \\
		Opt-IRL & 0.29 & 0.89 & 0.14 & 0.17 \\
		CIOC & 0.23 & 0.90 & 0.14 & 0.16 \\
		GCL &  3.73  &  46.70  &  1.53  & 4.69\\
		\hline
	\end{tabular}
% 	\vspace{-3pt}
\end{table}

\begin{table}[h!]
	\centering
	\caption{Generalization results of different IRL algorithms under the probabilistic metric.}
	\label{table:generalization_prob}
	\begin{tabular}{c|cc| cc}
		\hline
		&  {Seen NID} &  {Unseen NID} &  {Seen ID} &  {Unseen ID}\\
		\hline
		Ours &  -238.98  &  -399.85  & -515.97 & -571.60 \\
		Opt-IRL & -398.93 & -472.51 & -802.01 & -870.72 \\
		CIOC & -662.16 & -1153.74 & -595.27 & -621.13 \\
		GCL &  -1377.65  &  -3140.24  &  -1196.75  & -2898.64\\
		\hline
	\end{tabular}
% 	\vspace{-10pt}
\end{table}

\subsection{Computation Complexity}
We also compared the convergence speed of the four IRL algorithms. The time cost is summarized in \cref{table:time_cost}. We can see that the convergence speed of our method is significantly faster than that of {CIOC}, {Opt-IRL} and {GCL}. Such a superior convergence speed benefits from two reasons. First, the sampling method in {our} proposed approach is efficient (around 1 minute to generate all samples for the entire training set). Second, the sampling process is \emph{one-shot} in the algorithm through the training process. On the contrary, in each iteration, {Opt-IRL} needs to solve an optimization problem through gradient descent to find the optimal trajectory given the updated reward function. Such procedure can be extremely time-consuming, particularly when the planning time horizon is long. Thus, {Opt-IRL} might suffer from the scaling problem with long planning horizon and large training set. {GCL} also adopts a sampling-based method. However, it needs to re-generate all the samples in every training iteration, while our method only needs to generate all samples once. As for {CIOC}, the main computation load comes from the computation of gradient and hessian introduced via the Laplace approximation as shown in (\ref{eqa:laplace_approximation}). Besides, we also find the stability of the training performance for {CIOC} is quite sensitive to data noise and the selection of feature sets. Numerical computation issues, particularly for the hessian calculation, could happen and influence the learning performance in the presence of large data noise and/or miss-specified feature sets. %That is because of the Laplace approximation assumption in~\cite{levine2012continuous}. Real data is noisy and may not satisfy the local optimality assumption. This will cause numerical issues for the gradient $\frac{\partial R}{\partial \xi_{D}}$ and Hessian matrix $\frac{\partial^2 R}{\partial \xi_{D}^2}$ in Equation~\ref{eqa:laplace_approximation}. 
As a comparison, our proposed method is much less sensitive to either noise and feature selection, which is also a significant advantage.
\begin{table}[h]
	\centering
	\caption{The time cost of the three algorithms for both non-interactive and interactive scenarios. Results are in minutes}
	\label{table:time_cost}
	\begin{tabular}{c|cccc}
		\hline
		&  {Ours} &  {CIOC} &  {Opt-IRL} &  {GCL}\\
		\hline
		Non-interactive &  6  &  60  & 1800 & 40 \\
		Interactive &   5 &  90  &  1260  & 30\\
		\hline
	\end{tabular}
% 	\vspace{-10pt}
\end{table}

% \subsection{The Effect of Sample Re-Distribution}
% Here we investigate the effect of sample re-distribution step (in \cref{our_method:re-distribution}) for the learning performance. The experiment results with and without the re-distribution step for the non-interactive and interactive scenarios are shown in \cref{table:bin_sample_noninteractive} and \cref{table:bin_sample_interactive}, respectively. We can see that with the procedure of sample re-distribution, the proposed SMIRL algorithm can learn a better reward function in terms of the three categories of metrics: feature count deviations, MED and probabilistic metric of the ground-truth demonstrations in the test set. Most significant improvements are the results on ``Win Count'': the re-distribution of samples can help better evaluate the probabilities of ground-truth demonstrations. Such results are consistent with the conclusion from \cite{bobu2020less}, namely a uniformed distribution of samples via appropriate similarity function can help generate more accurate probabilistic predictions using the Boltzman noisily-rational model in (\ref{eq:max_entropy}).

\begin{table*}
\centering
\caption{Experiment results of the non-interactive scenario with and without the step of sample re-distribution\label{table:bin_sample_noninteractive}}
\begin{tabular}{c|cccc|c|c|c}
\hline
    &  a\_lon &  j\_lon &  v\_des &  a\_lat & MED & Win Count & Log Likelihood \\
\hline
    \makecell{w/ sample re-distribution} &      0.16$\pm$         0.12 &      0.20$\pm$         0.15 &      0.09$\pm$         0.04 &      0.09$\pm$         0.03 &     0.21$\pm$        0.06 & 33 &  -238.982\\
\hline
    \makecell{w/o sample re-distribution} &      0.18$\pm$         0.10 &      0.30$\pm$         0.16 &      0.12$\pm$         0.05 &      0.11$\pm$         0.04 &     0.26$\pm$        0.08 & 0 & -259.064\\
\hline
\end{tabular}
\end{table*}
\begin{table*}
\centering
\caption{Experiment results of the interactive scenario with and without the step of sample re-distribution}
\label{table:bin_sample_interactive}
\begin{tabular}{c|cccccc|c|c|c}
\hline
    &  a\_lon &  j\_lon &  a\_lat &  v\_des &  fut\_dis &  fut\_int\_dis &  MED & Win Count & Log Likelihood\\
\hline
    \makecell{w/ sample \\re-distribution} &      \makecell{0.14 $\pm$ \\  0.24} &  \makecell{0.53 $\pm$ \\  0.18} &   \makecell{0.19 $\pm$ \\       0.23} &   \makecell{0.032 $\pm$ \\  0.026} &    \makecell{0.012 $\pm$ \\   0.0074} &    \makecell{ 0.027 $\pm$  \\  0.044} &      \makecell{0.072 $\pm$  \\  0.043} & 76 & -515.965 \\
\hline
    \makecell{w/o sample \\re-distribution} &   \makecell{0.23 $\pm$ \\  0.53} &    \makecell{0.55 $\pm$ \\ 0.18} &   \makecell{0.19 $\pm$ \\ 0.23} &      \makecell{0.031 $\pm$ \\  0.028} &    \makecell{0.012 $\pm$ \\   0.0062} &     \makecell{0.027 $\pm$ \\  0.045} &  \makecell{0.067 $\pm$ \\  0.041} &  0 & -557.307 \\
\hline
\end{tabular}
\end{table*}

\subsection{The Effect of Sample Re-Distribution}
We investigated the effect of sample re-distribution step (in \cref{our_method:re-distribution}) for the learning performance. The experiment results with and without the re-distribution step for the non-interactive and interactive scenarios are shown in \cref{table:bin_sample_noninteractive} and \cref{table:bin_sample_interactive}, respectively. We can see that with the procedure of sample re-distribution, the proposed SMIRL algorithm can learn a better reward function in terms of the three categories of metrics: feature count deviations, MED and probabilistic metric of the ground-truth demonstrations in the test set. Most significant improvements are the results on ``Win Count'': the re-distribution of samples can help better evaluate the probabilities of ground-truth demonstrations. Such results are consistent with the conclusion from \cite{bobu2020less}, namely a uniformed distribution of samples via appropriate similarity function can help generate more accurate probabilistic predictions using the Boltzmann noisily-rational model in (\ref{eq:max_entropy}).

\vspace{6pt}
\section{Conclusion}
\label{summary}
In this paper, we proposed a sampling-based maximum entropy inverse reinforcement learning (IRL) algorithm in continuous domain to efficiently learn human driving behaviors. By explicitly leveraging prior knowledge on vehicle kinematics and motion planning, an efficient sampler was designed to estimate the intractable partition term when retrieving the reward function. Such benefits were verified by experiments on both non-interactive and interactive driving scenarios using the INTERACTION dataset. Comparing to the other three popular IRL algorithms, the proposed algorithm achieved better results in terms of both deterministic metrics such as feature count deviation and mean Euclidean distance, and probabilistic metrics such as the likelihood of demonstrations in the test set. Moreover, the proposed IRL algorithm shows better generalization ability and converges significantly faster than the other baseline methods.

Currently, the proposed IRL algorithm was designed specifically for mobile robot systems such as ground vehicles. In the future, extension to general robotic systems with higher dimensions such as robot manipulators will be considered. Moreover, the Euclidean distance metric used in the re-sampling step is not necessarily the best metric, and we will explore better metrics in future works.

%%%%%%%%%%%%%%%%%%%%%%%%%%%%%%%%%%%%%%%%%%%%%%%%%%%%%%%%%%%%%%%%%%%%%%%%%%%%%%%%

%%%%%%%%%%%%%%%%%%%%%%%%%%%%%%%%%%%%%%%%%%%%%%%%%%%%%%%%%%%%%%%%%%%%%%%%%%%%%%%%

%%%%%%%%%%%%%%%%%%%%%%%%%%%%%%%%%%%%%%%%%%%%%%%%%%%%%%%%%%%%%%%%%%%%%%%%%%%%%%%%
\section*{ACKNOWLEDGMENT}
The authors would like to thank students Zhaoting Li, Songyuan Zhang, Fan Yang and Chenran Li for their help in implementing the sampling algorithm in \cref{our_method:sampler}, and thank Fenglong Song for generating all reference paths based on the high-definition maps.

%%%%%%%%%%%%%%%%%%%%%%%%%%%%%%%%%%%%%%%%%%%%%%%%%%%%%%%%%%%%%%%%%%%%%%%%%%%%%%%%
\bibliographystyle{IEEEtran}
\bibliography{references}

\end{document}